\documentclass[10pt]{article}

\usepackage[utf8]{inputenc}
\usepackage[T1]{fontenc}
\usepackage{epsfig}
\usepackage{subfigure}
\usepackage{graphicx}
\usepackage{algorithm}
\usepackage{algorithmic}
\usepackage{mathrsfs}
\usepackage{amsfonts}
\usepackage{float}
\usepackage{amssymb,amsmath}
\usepackage{indentfirst}
\usepackage{bm}
\usepackage{dsfont}
\usepackage{makecell}
\usepackage{theorem}
\usepackage{multirow}
\usepackage{array}
\usepackage{multirow}
\usepackage{subfigure}
\usepackage{longtable}
\usepackage{threeparttable}
\usepackage[table]{xcolor}
\usepackage{booktabs}
\usepackage{todonotes}

\newif\iftwoc
\twocfalse

\newif\ifpfig
\pfigtrue

\iftwoc

\makeatletter
\def\@normalsize{\@setsize\normalsize{12pt}\xpt\@xpt
\abovedisplayskip 12pt plus3pt minus7pt\belowdisplayskip \abovedisplayskip
\abovedisplayshortskip \z@ plus3pt\belowdisplayshortskip 6.5pt plus3.5pt
minus3pt\let\@listi\@listI}

\def\subsize{\@setsize\subsize{12pt}\xipt\@xipt}

\def\section{\@startsection {section}{1}{\z@}{24pt plus 2pt minus 2pt}
{12pt plus 2pt minus 2pt}{\large\bf}}

\def\subsection{\@startsection {subsection}{2}{\z@}{12pt plus 2pt minus 2pt}
{12pt plus 2pt minus 2pt}{\subsize\bf}}
\makeatother

\else

\renewcommand{\baselinestretch}{1.2}

\fi

\newlength{\wone}
\newcommand{\eq}{\begin{equation}}
\newcommand{\en}{\end{equation}}

\def\DIC{{\hbox{\rm\kern.2em\raise.36ex%
\hbox{$\scriptstyle |$}\kern-.4em C}}}
\def\SIC{{\hbox{\scriptsize\rm\kern.2em\raise.4ex%
\hbox{$\scriptscriptstyle |$}\kern-.4em C}}}

\def\DIQ{{\hbox{\rm\kern.2em\raise.4ex%
\hbox{$\scriptstyle |$}\kern-.4em Q}}}
\def\SIQ{{\hbox{\scriptsize\rm\kern.2em\raise.4ex%
\hbox{$\scriptscriptstyle |$}\kern-.4em Q}}}

\def\DZZ{{\hbox{\sf Z\kern-.41em Z}}}
\def\SZZ{{\hbox{\scriptsize\sf Z\kern-.41em Z}}}

\begin{document}
% don't want date printed
\date{}
\title{\Large\bf PolySmart @ TRECVid 2024 Video Captioning (VTT)}
\author{
	Jiaxin Wu$^\dagger$, Wengyu Zhang$^\dagger$, Xiao-Yong Wei$^{\star\dagger}$, Qing Li$^\dagger$ 
    \vspace{0.08in}
   \\ {\em $^\dagger$Department of Computing, The Hong Kong Polytechnic University}
     \\    {\em $^\star$Department of Computer Science, Sichuan University}
     \vspace{0.08in}
   \\ nikki-jiaxin.wu@polyu.edu.hk, wengyu.zhang@connect.polyu.hk\\x1wei@polyu.edu.hk, qing-prof.li@polyu.edu.hk\\
}
\maketitle

% ------------------------------
% % Begin abstract
% % ------------------------------
\section*{\centering Abstract}
\indent In this paper, we present our methods and results for the Video-To-Text (VTT) task at TRECVid 2024~\cite{2023trecvidawad}, 
exploring the capabilities of Vision-Language Models (VLMs) like LLaVA and LLaVA-NeXT-Video in generating natural language descriptions for video content.
We investigate the impact of fine-tuning VLMs on VTT datasets to enhance description accuracy, contextual relevance, and linguistic consistency.
Our analysis reveals that fine-tuning substantially improves the model's ability to produce more detailed and domain-aligned text, bridging the gap between generic VLM tasks and the specialized needs of VTT.
Experimental results demonstrate that our fine-tuned model outperforms baseline VLMs across various evaluation metrics,
underscoring the importance of domain-specific tuning for complex VTT tasks.

% -------------------------------
% Begin body
% -------------------------------

% -------------------------------
% Begin introduction
% -------------------------------
\section{Video-To-Text (VTT)}
The Video-to-Text (VTT) task poses the challenge of generating concise, accurate natural language descriptions for video content, 
which is a complex vision-language task critical in domains like accessibility, content retrieval, and human-computer interaction. 
Similarly to text-video retrieval \cite{chongwahngo2005trecvid,chongwahngo2008trecvid,chongwahngo2010trecvid}, the VTT task requires integrating visual information with language processing to have a good understanding of video content \cite{vireo2023,vireo2022,vireo2021, improvedITV, wu2023ITV}. 
With advancements in Large Language Model(LLM) and Vision Language Model (VLM) like LLaMA~\cite{touvron2023llama} and LLaVA~\cite{liu2024visual}, researchers have demonstrated the ability of these models to understand visual and textual information \cite{VBS2025}.  

Therefore, we consider leveraging the power of VLM models in the VTT task for better text description generation. 
Specifically, we utilize the LLaVA~\cite{liu2024visual} and LLaVA-NeXT-Video~\cite{zhang2024video} model for the VTT task.
VLM has been pre-trained on large amounts of visual-textual data and fine-tuned with instructions for specific tasks,
such as video understanding, video question answering, and video captioning.
We further fine-tune the VLM on a large amount of VTT video-text pairs, aiming to enable the model to specialize in the Video-To-Text task.

% -------------------------------
% Begin Method
% -------------------------------
\section{Method}

To generate description text from videos, the following three methods are applied.

\begin{figure*}[http]
    \centering
    \includegraphics[width=0.6\linewidth]{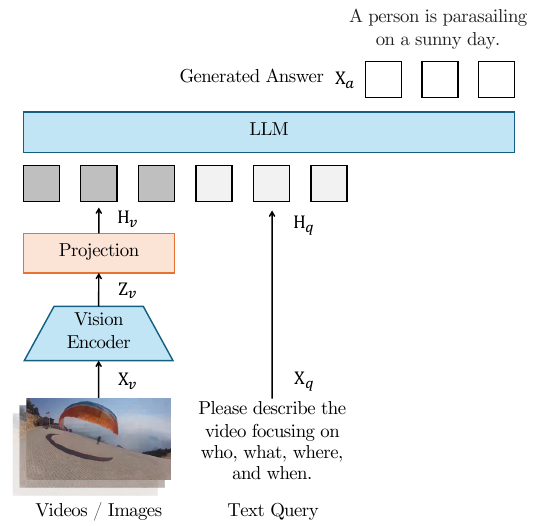}
    \caption{Vision Language Model Framework.}
    \label{fig:llava_pipeline}
\end{figure*}

\subsection{Generating Text Description by Vision Language Models}

Taking advantage of pre-trained capabilities of Large Language Models and Visual Encoder as well as visual instruction tuning, 
VLM gains considerable prior knowledge on video understanding tasks~\cite{zhang2024video}.
The model framework is shown in Figure~\ref{fig:llava_pipeline}.
A backbone LLM is used for visual understanding and text generation.
A visual encoder transforms input visual information $\mathbf{X}_v$ into visual embeddings $\mathbf{Z}_v$.
In order to align the embedding space, an MLP~\cite{liu2024improved} block projects visual embeddings $\mathbf{Z}_v$ into LLM's token embeddings $\mathbf{H}_v$.
The probability of the target answers $\mathbf{X}_a$ to question$\mathbf{X}_q$ and video $\mathbf{X}_v$ is defined as

$$
p(\mathbf{X}_\mathrm{a} \mid \mathbf{X}_\mathrm{v}, \mathbf{X}_\mathrm{q}) = \prod_{i=1}^{L} p(x_i \mid \mathbf{X}_\mathrm{v}, \mathbf{X}_{\mathrm{a}, <i}, \mathbf{X}_{\mathrm{q}, <i}).
$$

The training process of VLMs typically consists of two stages,
\begin{itemize}
    \item Vision-Language Alignment: to align the visual embeddings with LLM's text embeddings, as LLM's visual ability acquisition, and
    \item Visual Instruction Tuning: to give LLM abilities to complete different kinds of visual taskings, such as video captioning, video short question answering, and video multiple-choice question answering.
\end{itemize}

\textbf{LLaVA}

We use the LLaVA-v1.5-7b model\footnote{https://huggingface.co/liuhaotian/llava-v1.5-7b} and follow the official instructions\footnote{https://github.com/haotian-liu/LLaVA/tree/main?tab=readme-ov-file\#quick-start-with-huggingface} to perform the captioning on VTT24 videos. 
We extract the middle frame of each video as the image input of the model. The text query we used is

\texttt{"Please write a description of this video frame (around 20-30 words), focusing on\\Who, What, Where, and When."}.

\vspace{1em}
\textbf{LLaVA-NeXT-Video}

We use the LLaVA-NeXT-Video-7B-DPO model\footnote{https://huggingface.co/lmms-lab/LLaVA-NeXT-Video-7B-DPO} and follow the official instructions\footnote{https://github.com/LLaVA-VL/LLaVA-NeXT/blob/main/docs/LLaVA-NeXT-Video.md} to perform the captioning on VTT24 videos.
The entire video is used as the video input of the model. The prompt we used is:

\texttt{"Please provide a detailed description of the video, focusing on the main subjects,\\their actions, the background scenes."}.

\subsection{Fine-tuning Vision Language Models on VTT Task}

The capabilities of vision language models (VLMs) have been investigated on many tasks~\cite{zhu2023minigpt,li2025llama,ataallah2024minigpt4}. However, the vanilla VLMs are typically fine-tuned on specific dataset and instructions,
They have noticeable domain gaps on VTT tasks, such as the length of generated text, graininess of description, and wording.
Therefore, we decide to fine-tune the VLMs specifically on the VTT dataset for better text description generation.
We choose LLaVA~\cite{liu2024llava} as our VLM,
and follows official instructions\footnote{https://github.com/haotian-liu/LLaVA/blob/main/docs/Finetune\_Custom\_Data.md} to fine-tune the model on VTT dataset. We do not fine-tune the LLaVA-NeXT-Video model since no official fine-tuning code is available.

We collect video-text pairs from VTT16-VTT23 datasets~\cite{rossetto2019v3c}.
For each video, we extract frames for each of the 5 frames, resulting in $699683$ frame-text pairs for fine-tuning.
The fine-tuning text query we used is

\texttt{"Please write a description of this video frame (around 20-30 words), focusing on\\Who, What, Where, and When."}.

% -------------------------------
% Begin Results analysis
% -------------------------------
\section{Results analysis}

\begin{table}[h!]
\caption{Performance comparison among Fine-tuned LLaVA (LV-FT), Vanilla LLaVA (LV) and Vanilla LLaVA-NeXT-Video (LV-V). The best performances are in bold. Rob.: Robustness.}
\label{table:score}
    \centering
    \footnotesize % \scriptsize \footnotesize /small
    \begin{tabular}{ccc*{10}c}
        \toprule
        \textbf{Run} & \textbf{Model} & \textbf{Task} & \textbf{BL~$\uparrow$} & \textbf{ME~$\uparrow$} & \textbf{CI~$\uparrow$} & \textbf{CD~$\uparrow$} & \textbf{SP~$\uparrow$} & \textbf{S1~$\uparrow$} & \textbf{S2~$\uparrow$} & \textbf{S3~$\uparrow$} & \textbf{S4~$\uparrow$} & \textbf{S5~$\uparrow$} \\
        \midrule
        1 & LV-FT & \multirow{3}{*}{Main} & \textbf{0.128} & \textbf{0.379} & \textbf{0.712} & \textbf{0.427} & 0.149 & 0.448 & 0.446 & \textbf{0.482} & \textbf{0.446} & \textbf{0.452} \\
        3 & LV &                                       & 0.101 & 0.324 & 0.637 & 0.323 & 0.114 & 0.432 & 0.438 & 0.458 & 0.418 & 0.431 \\
        4 & LV-V &                                       & 0.027 & 0.286 & 0.511 & 0.015 & \textbf{0.156} & \textbf{0.459} & \textbf{0.447} & 0.478 & 0.432 & 0.451 \\
        \midrule
        1 & LV-FT & \multirow{3}{*}{Rob.} & \textbf{0.131} & \textbf{0.377} & \textbf{0.715} & \textbf{0.443} & 0.147 & 0.444 & \textbf{0.445} & 0.467 & \textbf{0.444} & 0.444 \\
        3 & LV &                                          & 0.105 & 0.318 & 0.634 & 0.321 & 0.113 & 0.432 & 0.438 & 0.448 & 0.414 & 0.433 \\
        4 & LV-V &                                          & 0.027 & 0.293 & 0.490 & 0.016 & \textbf{0.158} & \textbf{0.456} & 0.441 & \textbf{0.474} & 0.428 & \textbf{0.445} \\
        \bottomrule
    \end{tabular}
\end{table}

The evaluation result of 3 proposed methods on VTT24 is shown in~Table \ref{table:score}. We report the BLEU (BL), METEOR (ME), CIDEr (CI), CIDEr-D (CD), SPICE (SP) and STS1-5 (S1-S5) scores, aligning with~\cite{1178722}.

\subsection{The Capability of Pre-trained VLM}

We explore the inherent capabilities of the pre-trained Visual Language Model (VLM) without fine-tuning. We focus on evaluating the model’s performance across different tasks when presented with video or frames as inputs.

\subsubsection{Video as Input}

When video data is directly input to the model, we observe notable performance differences across the three model variants.
Table~\ref{table:score} shows that the Vanilla LLaVA-NeXT-Video (LV-V) generally lags in metrics such as BLEU (BL), METEOR (ME), and CIDEr (CI) compared to the fine-tuned and vanilla LLaVA (LV-FT and LV).
This suggests that the video variant may not fully leverage sequential data without additional training.
Notably, LV-V achieves a higher SPICE (SP) score, 
indicating a better understanding of semantic relationships in certain scenarios.
However, its performance inconsistency implies limited generalization when processing raw video as input.

\subsubsection{Frames as Input}

In contrast, when input is fed with frames, the vanilla LLaVA (LV) model performs more robustly than the LLaVA-NeXT-Video (LV-V) model across the board. 
The results in Table~\ref{table:score} reveal that LV achieves higher scores across most metrics, including BLEU, METEOR, and CIDEr. 
This indicates that frame-based inputs allow LV to capture detailed content more effectively than LV-V, which directly processes video inputs.
The consistent advantage of frame-by-frame input suggests that LV better leverages individual frame details to understand and align with ground-truth captions.
This approach likely aids the model in capturing nuances that might be lost when processing continuous video sequences as a single input.

\subsection{Impact of Fine-tuning}

The impact of fine-tuning is evident when comparing LV-FT to the other two models across both main and robustness (Rob.) tasks.
LV-FT achieves the highest scores in almost all metrics in the main task, indicating that fine-tuning enhances both syntactic (BLEU, METEOR) and semantic (CIDEr, CIDEr-D) understanding.
Specifically, LV-FT’s substantial improvement in CIDEr-D and BLEU scores implies that fine-tuning has refined its ability to capture detailed and relevant content, particularly in challenging scenarios.

Among the 300 queries in VTT24, 171 ($57\%$) showed an improvement in CIDEr scores with the fine-tuned model (LV-FT), compared to the vanilla model (LV).
On average, these improved queries achieved a mean gain of 0.356, with the largest observed increase reaching 1.789 in CIDEr score.

\begin{figure*}[http]
    \centering
    \begin{minipage}[b]{0.46\linewidth}
        % \centering
        \includegraphics[width=\linewidth]{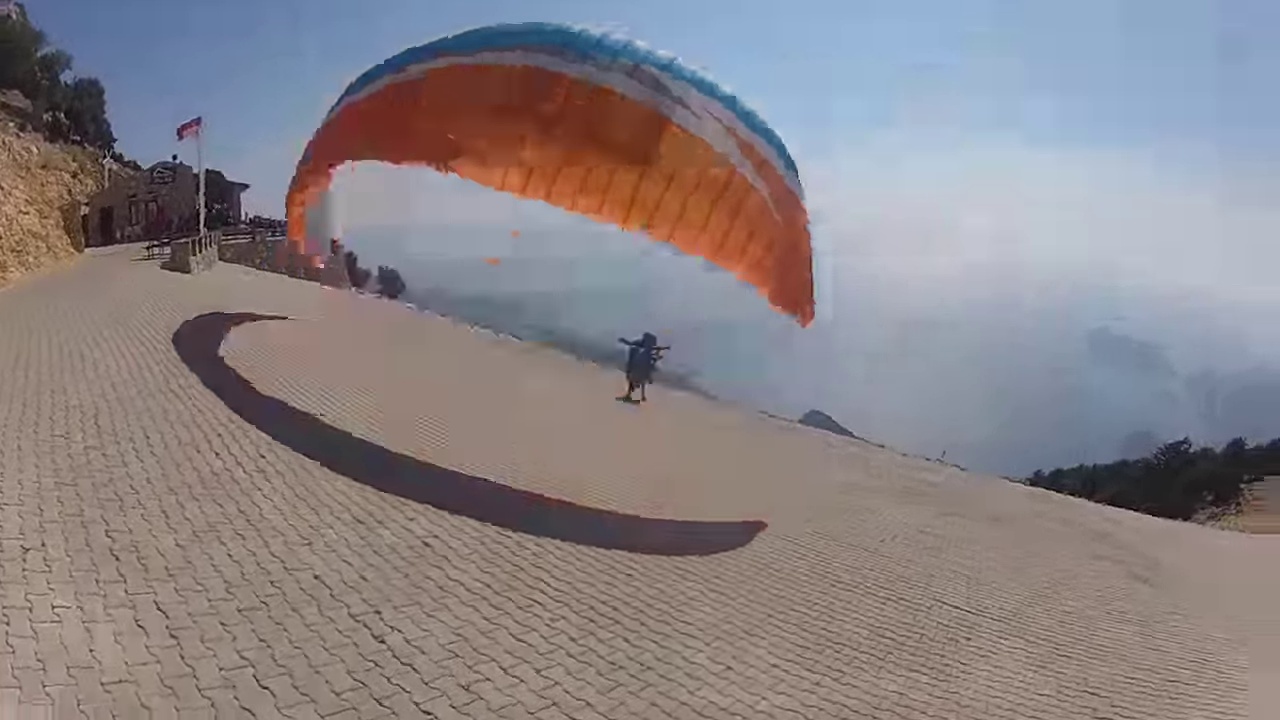}
        \\ \textbf{LV}: A person is parasailing over a \textcolor{red}{beach}. \\ \textbf{LV-FT}: A person is parasailing on a \textcolor{blue}{sunny day}.\\ \\
    \end{minipage}
    \hspace{0.02\linewidth}
    \begin{minipage}[b]{0.46\linewidth}
        % \centering
        \includegraphics[width=\linewidth]{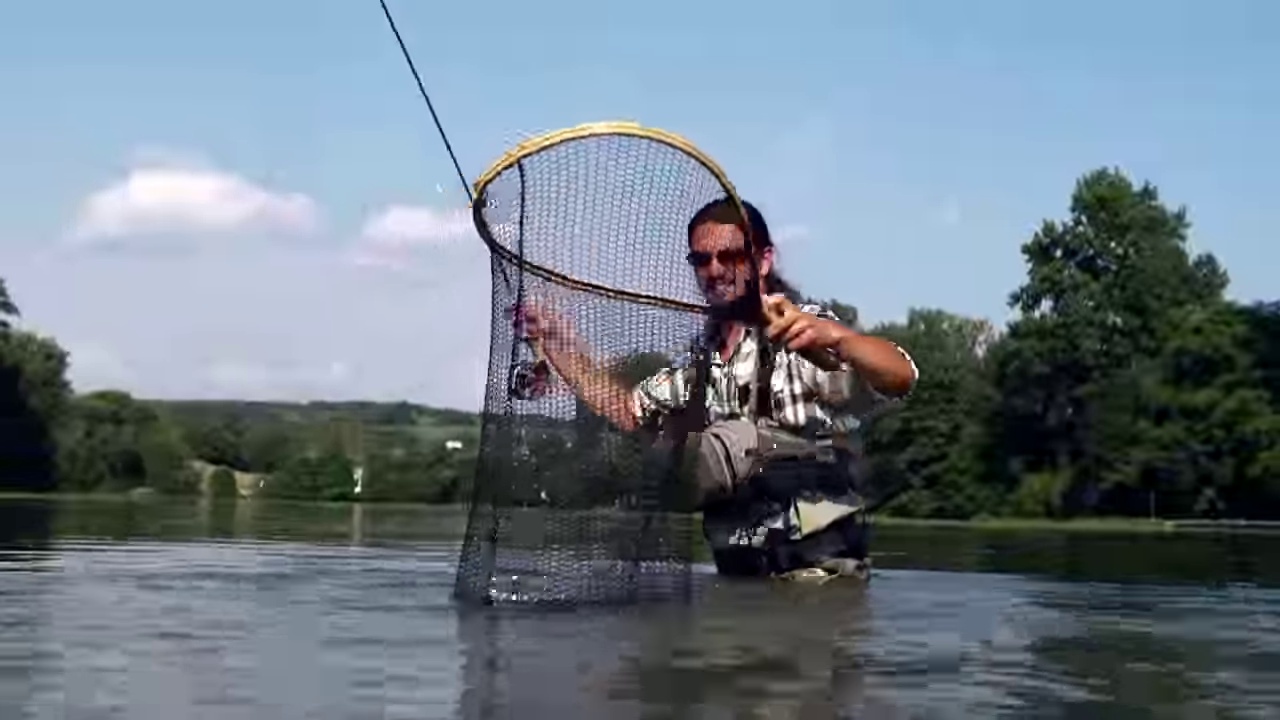}
        \\ \textbf{LV}: A man is holding a fishing net in the water. \\ \textbf{LV-FT}: A man in a \textcolor{blue}{khaki shirt} and a fishing net stands in the water on a sunny day. \\
    \end{minipage}

    \begin{minipage}[b]{0.46\linewidth}
        % \centering
        \includegraphics[width=\linewidth]{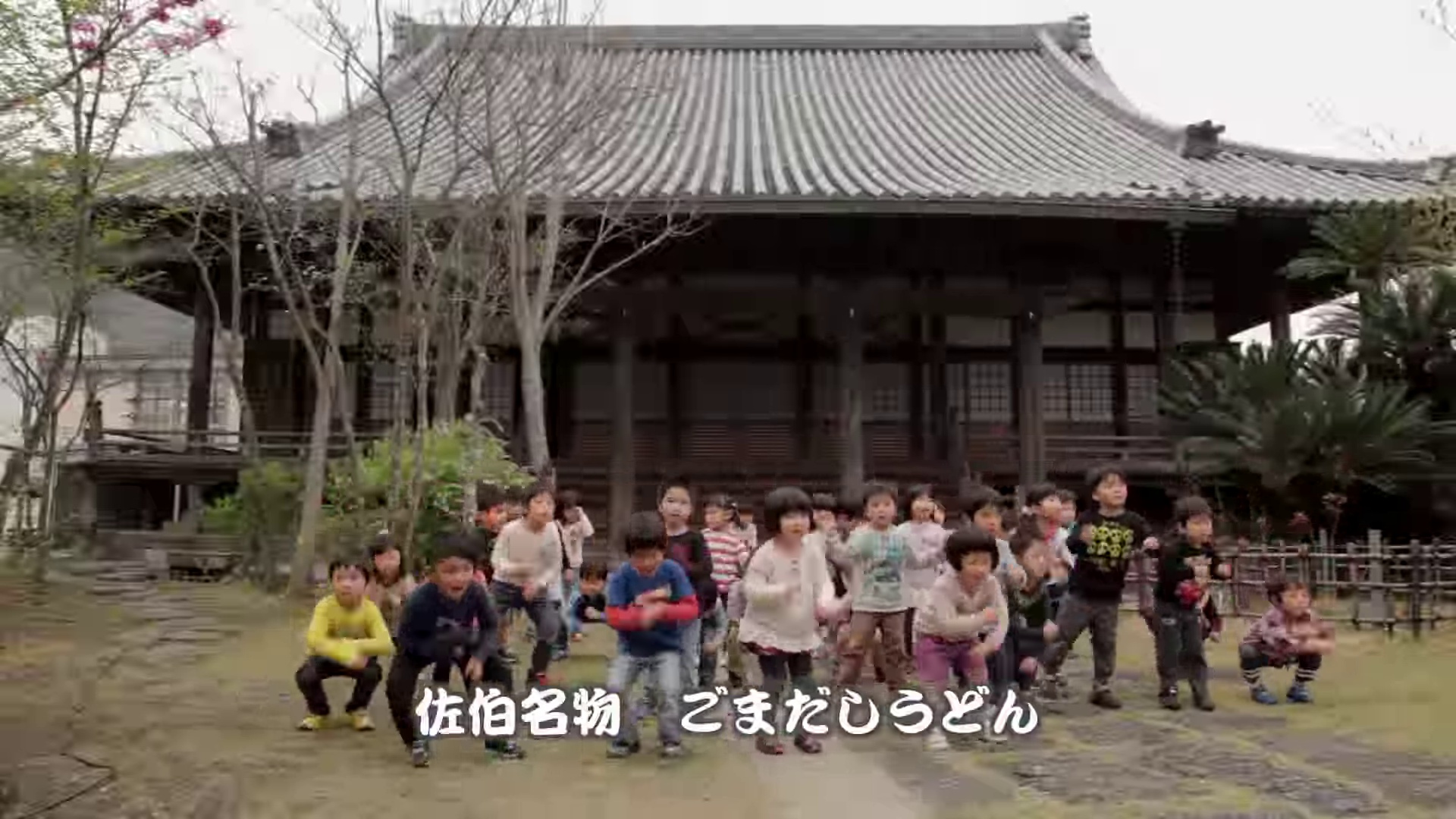}
        \\ \textbf{LV}: A group of children standing in front of a building. \\ \textbf{LV-FT}: A group of \textcolor{blue}{Asian} children are standing in front of a building and bowing outside on a sunny day.
    \end{minipage}
    \hspace{0.02\linewidth}
    \begin{minipage}[b]{0.46\linewidth}
        % \centering
        \includegraphics[width=\linewidth]{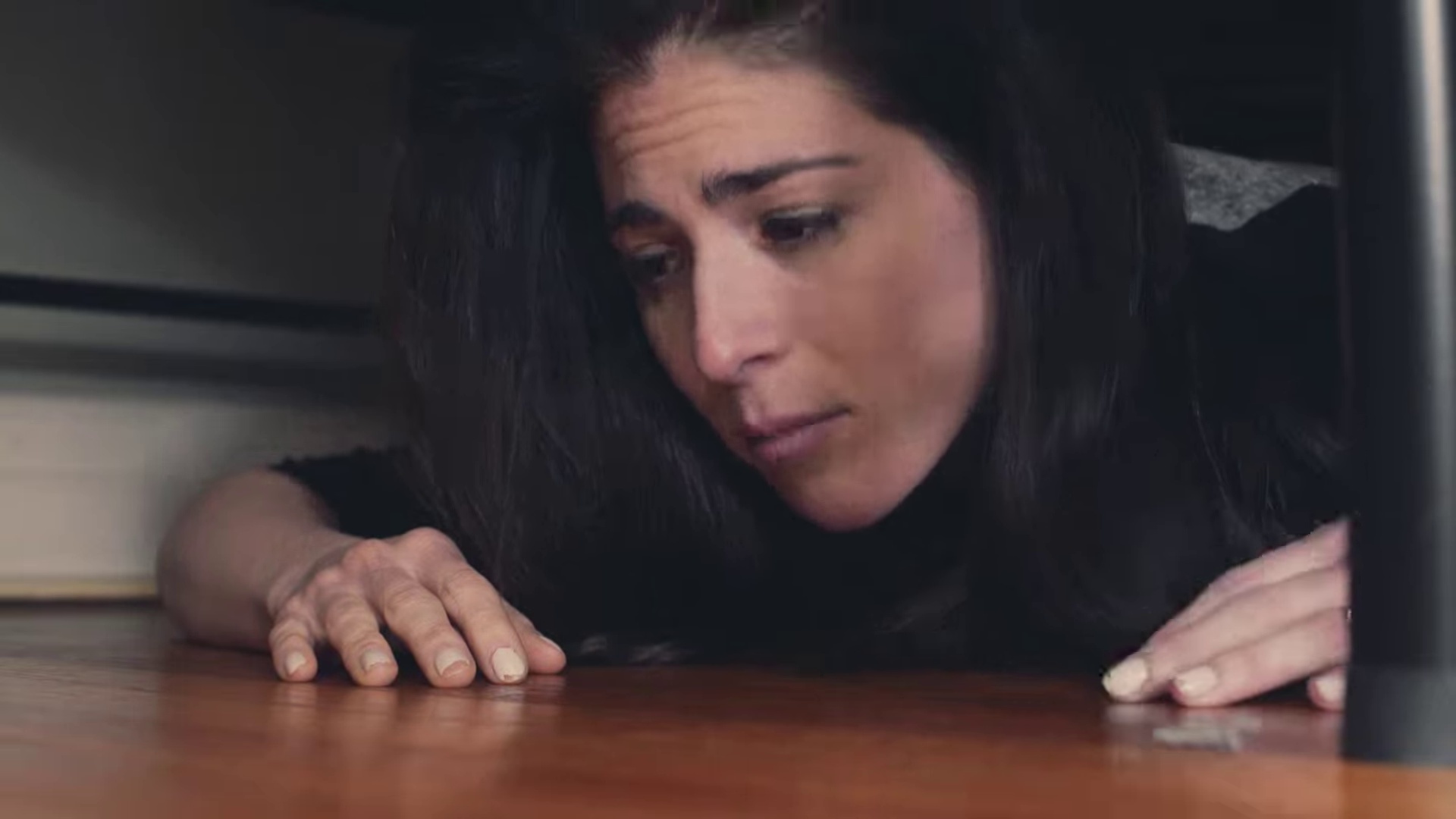}
        \\ \textbf{LV}: A woman with long hair and a black shirt is looking down at the floor. \\ \textbf{LV-FT}: A woman with long dark hair is \textcolor{blue}{lying} on the floor and looking at something in front of her. 
    \end{minipage}
    \caption{Case study among Vanilla LLaVA (LV) and Fine-tuned LLaVA (LV-FT)}
    % \vspace{-1em}
    \label{fig:case_study}
\end{figure*}

\subsection{Discussions}

\subsubsection{Fine-tuning Brings VLMs Detailed Description}

We discuss the effects of fine-tuning on Vision-Language Models (VLMs),
specifically illustrated through the differences in descriptions generated by Vanilla LLaVA (LV) and Fine-Tuned LLaVA (LV-FT) models.
Fine-tuning appears to enhance the model's attention to detail, context, and specificity in its generated descriptions.

In Figure \ref{fig:case_study}, we observe that the LV model typically provides a general description, focusing on the primary subject and basic actions or objects.
However, the fine-tuned LV-FT model incorporates additional contextual information and descriptive details, as seen across all examples:

\begin{itemize}
    \item In the first image, LV incorrectly describes the ground as "beach". LV-FT, however, refines this by adding "on a sunny day", enriching the setting and potentially suggesting the mood or conditions.
    \item Similarly, in the second image, LV's description "a man holding a fishing net in the water" is accurate yet lacks specificity. LV-FT adds that the man is in a "khaki shirt" and that it's a "sunny day," thus providing more visual clues.
    \item The third image illustrates a significant enhancement in recognizing demographic context: LV mentions "a group of children standing in front of a building," while LV-FT specifies "Asian children bowing outside on a sunny day," adding cultural and situational context, which could improve performance in cultural or geographic datasets.
    \item Lastly, in the fourth example, LV's description is correct but lacks details about her pose. LV-FT adds that she is "lying on the floor and looking at something in front of her," capturing a more complete view of her posture and engagement with the environment.
\end{itemize}

Overall, these examples demonstrate that fine-tuning enables VLMs to capture richer contextual details and subtle variations in image,
which could enhance performance in applications requiring a nuanced understanding of visual content.
Fine-tuning not only reinforces the model's ability to describe the primary elements but also to interpret contextual cues, like clothing color, setting, or implied cultural details.
These findings suggest that fine-tuning is an essential step for optimizing VLMs for real-world applications that rely on precise and contextually aware descriptions.

\subsubsection{Fine-tuning Aligns VLMs' Responses with VTT Tasks}

\begin{figure*}[http]
    \centering
    \begin{minipage}[b]{0.3\linewidth}
        \centering
        \includegraphics[width=\linewidth]{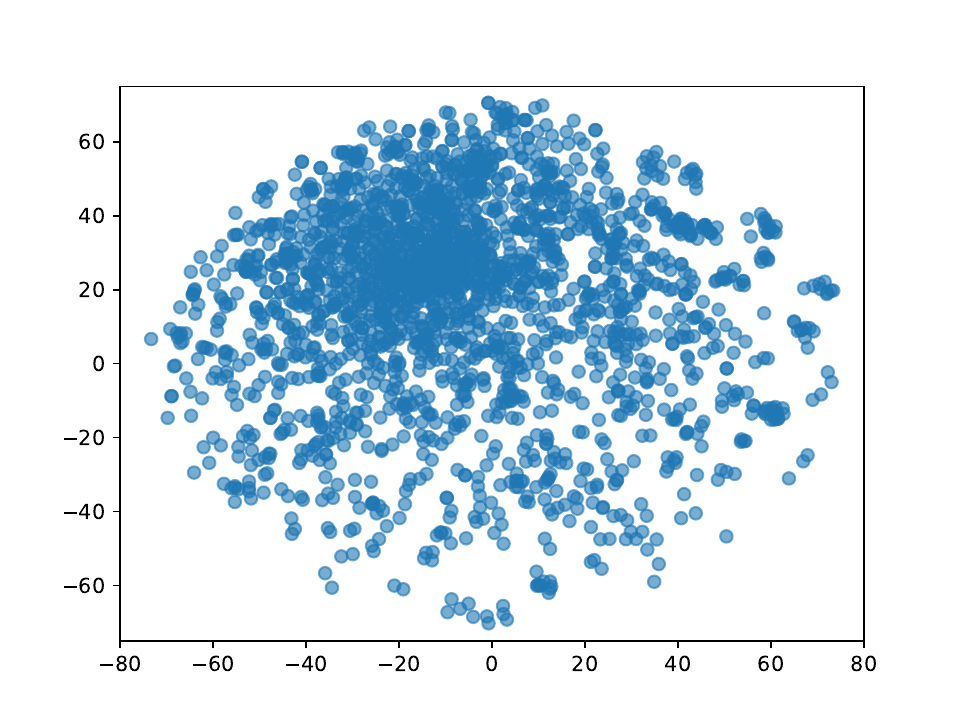}
        \\ (a) \textbf{VTT16-23} \\dataset text
    \end{minipage}
    \hspace{0.02\linewidth}
    \begin{minipage}[b]{0.3\linewidth}
        \centering
        \includegraphics[width=\linewidth]{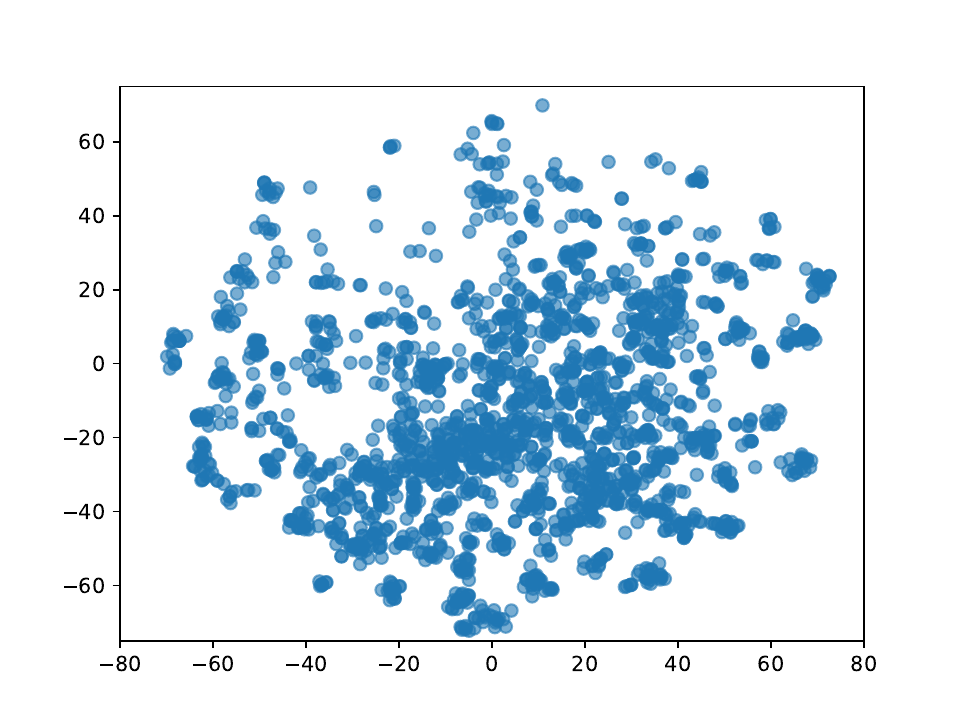}
        (b) \textbf{Vanilla} LLaVA (LV) generated text 
    \end{minipage}
    \hspace{0.02\linewidth}
    \begin{minipage}[b]{0.3\linewidth}
        \centering
        \includegraphics[width=\linewidth]{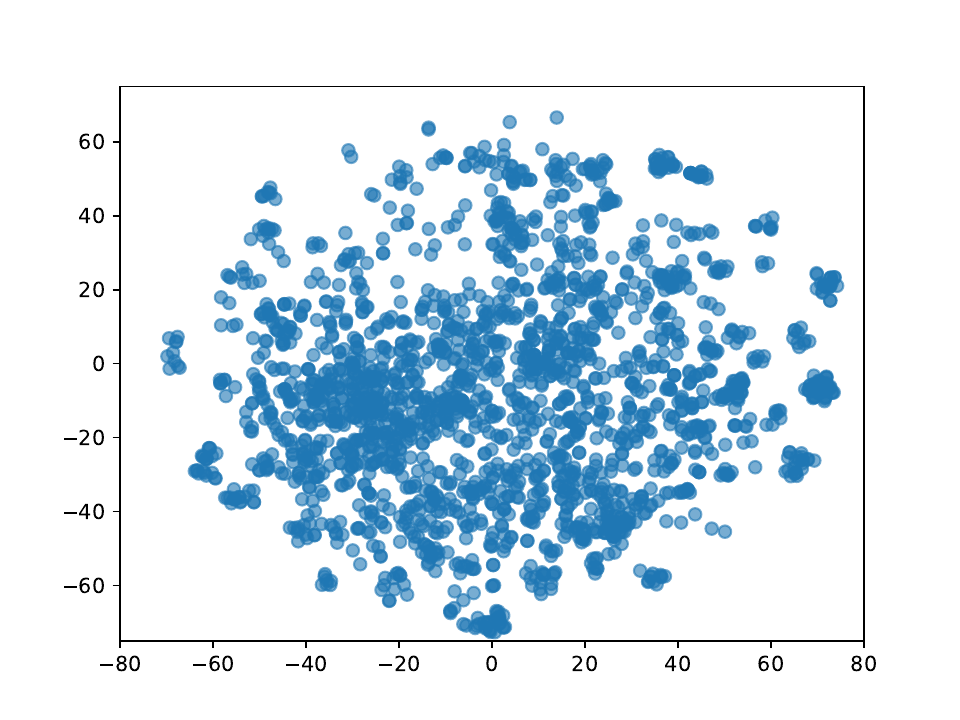}
        (c) \textbf{Fine-tuned} LLaVA (LV-FT) generated text
    \end{minipage}
    % \hspace{0.02\linewidth}
    
    \caption{Comparison among Text Embedding t-SNE Distributions.}
    % \vspace{-1em}
    \label{fig:text_tsne}
\end{figure*}

% We believe fine-tuning narrows domain gaps between vanilla VLM tasks and VTT tasks, such as the graininess of description and wording. In Figure~\ref{fig:text_tsne}, we visualize the text embedding distribution of (a) VTT16-23 text description, (b) Vanilla LLaVA (LV), and (c) Fine-tuned LLaVA (LV-FT). Notably, LV-FT generated text has a more even distribution than LV, where LV-FT has been fine-tuned on the frame-text pairs extracted from VTT16-23 dataset. Therefore, the distribution of LV-FT generated text shifts towards the distribution of text in VTT16-23 dataset, resulting in a text expression more suitable for the VTT task.

Fine-tuning aligns Vision-Language Models (VLMs) with the requirements of VTT tasks by reducing the domain gap,
not only improving description detail but also enhancing text style coherence.
In Figure~\ref{fig:text_tsne}, the t-SNE visualizations highlight that fine-tuning on VTT16-23 frame-text pairs shifts the distribution of LV-FT's generated text closer to the target dataset (VTT16-23) than that of Vanilla LLaVA (LV). 

This shift indicates several improvements:

\begin{itemize}
    \item \textbf{Domain Alignment}: Fine-tuning narrows the gap between generic VLM outputs and the specificity required in VTT, resulting in descriptions that better match the visual and contextual details needed for effective task execution.
    \item \textbf{Linguistic Consistency}: LV-FT’s text adopts a style closer to that of VTT16-23 dataset, ensuring descriptions are more cohesive and consistent with domain language patterns.
    \item \textbf{Enhanced Versatility}: Compared to LV, LV-FT generates more granular and contextually accurate descriptions. The balanced distribution in LV-FT indicates its adaptability across diverse scenes, improving robustness in VTT applications.
\end{itemize}

These results highlight the importance of fine-tuning in optimizing VLMs for specialized tasks, enhancing both descriptive quality and consistency in VTT settings.

% -------------------------------
% Begin Conclusion
% -------------------------------
\section{Conclusion}

This study demonstrates the effectiveness of fine-tuning Vision-Language Models (VLMs) for the Video-To-Text (VTT) task,
highlighting significant improvements in descriptive detail, contextual accuracy, and linguistic alignment.
Our experiments show that while pre-trained VLMs exhibit inherent video understanding capabilities,
fine-tuning on a VTT-specific dataset enhances their performance across multiple metrics.
The fine-tuned LLaVA model (LV-FT) consistently outperforms the vanilla and video-specific models,
achieving higher scores in both syntactic and semantic metrics.
These findings suggest that VLMs can be optimized for domain-specific tasks by adapting to dataset characteristics,
enabling more accurate and context-aware video descriptions.
Future work could extend these findings by exploring additional fine-tuning strategies and evaluating their impact on other complex vision-language tasks.

\section{Acknowledgments}
This research project is supported by the National Natural Science Foundation of China (Grant No.: 62372314). 

\bibliographystyle{IEEEtran}
\bibliography{trec}

\begin{thebibliography}{10}
\providecommand{\url}[1]{#1}
\csname url@samestyle\endcsname
\providecommand{\newblock}{\relax}
\providecommand{\bibinfo}[2]{#2}
\providecommand{\BIBentrySTDinterwordspacing}{\spaceskip=0pt\relax}
\providecommand{\BIBentryALTinterwordstretchfactor}{4}
\providecommand{\BIBentryALTinterwordspacing}{\spaceskip=\fontdimen2\font plus
\BIBentryALTinterwordstretchfactor\fontdimen3\font minus \fontdimen4\font\relax}
\providecommand{\BIBforeignlanguage}[2]{{%
\expandafter\ifx\csname l@#1\endcsname\relax
\typeout{** WARNING: IEEEtran.bst: No hyphenation pattern has been}%
\typeout{** loaded for the language `#1'. Using the pattern for}%
\typeout{** the default language instead.}%
\else
\language=\csname l@#1\endcsname
\fi
#2}}
\providecommand{\BIBdecl}{\relax}
\BIBdecl

\bibitem{2023trecvidawad}
G.~Awad, K.~Curtis, A.~A. Butt, J.~Fiscus, A.~Godil, Y.~Lee, A.~Delgado, E.~Godard, L.~Diduch, Y.~Graham, , and G.~Quénot, ``Trecvid 2023 - a series of evaluation tracks in video understanding,'' in \emph{Proceedings of TRECVID 2023}.\hskip 1em plus 0.5em minus 0.4em\relax NIST, USA, 2023.

\bibitem{chongwahngo2005trecvid}
C.-W. Ngo, Z.~Pan, X.~Wei, X.~Wu, H.-K. Tan, and W.~Zhao, ``Motion driven approaches to shot boundary detection, low-level feature extraction and bbc rushes characterization at {TRECV}id 2005,'' in \emph{TRECVID}, 2005.

\bibitem{chongwahngo2008trecvid}
C.-W. Ngo, Y.-G. Jiang, X.-Y. Wei, W.~Zhao, F.~Wang, X.~Wu, and H.-K. Tan, ``Beyond semantic search: What you observe may not be what you think,'' in \emph{IEEE Computer Society}, 2008.

\bibitem{chongwahngo2010trecvid}
C.-W. Ngo, S.-A. Zhu, H.-K. Tan, W.-L. Zhao, and X.-Y. Wei, ``{VIREO} at {TREC}v{ID} 2010: Semantic indexing, known-item search, and content-based copy detection,'' in \emph{TRECVID}, 2010.

\bibitem{vireo2023}
J.~Wu, Z.~Ma, C.-W. Ngo, and S.-H. Zhong, ``{VIREO}@{TRECV}id 2023: Ad-hoc video search,'' in \emph{In NIST TRECVID Workshop}, 2023.

\bibitem{vireo2022}
J.~Wu, Z.~Ma, and C.-W. Ngo, ``{VIREO}@{TRECV}id 2022: Ad-hoc video search,'' in \emph{In NIST TRECVID Workshop}, 2022.

\bibitem{vireo2021}
J.~Wu, Z.~Hou, Z.~Ma, and C.-W. Ngo, ``{VIREO}@{TRECV}id 2021: Ad-hoc video search,'' in \emph{In NIST TRECVID Workshop}, 2021.

\bibitem{improvedITV}
J.~Wu, C.~wah Ngo, and W.-K. Chan, ``Improving interpretable embeddings for ad-hoc video search with generative captions and multi-word concept bank,'' in \emph{Proceedings of the ACM on International Conference on Multimedia Retrieval}, 2024, pp. 1--10.

\bibitem{wu2023ITV}
J.~Wu, C.-W. Ngo, W.-K. Chan, and Z.~Hou, ``{(Un)}likelihood training for interpretable embedding,'' in \emph{ACM Transactions on Information Systems}, 2023.

\bibitem{touvron2023llama}
H.~Touvron, T.~Lavril, G.~Izacard, X.~Martinet, M.-A. Lachaux, T.~Lacroix, B.~Rozi{\`e}re, N.~Goyal, E.~Hambro, F.~Azhar \emph{et~al.}, ``Llama: Open and efficient foundation language models,'' \emph{arXiv preprint arXiv:2302.13971}, 2023.

\bibitem{liu2024visual}
H.~Liu, C.~Li, Q.~Wu, and Y.~J. Lee, ``Visual instruction tuning,'' \emph{Advances in neural information processing systems}, vol.~36, 2024.

\bibitem{VBS2025}
Y.-T. Cheng, J.~Wu, Z.~Ma, J.~He, X.-Y. Wei, and C.-W. Ngo, ``Interactive video search with multi-modal llm video captioning,'' in \emph{Proceedings of the International Conference on Multimedia Modelling}, 2025, pp. 1--8.

\bibitem{zhang2024video}
Y.~Zhang, J.~Wu, W.~Li, B.~Li, Z.~Ma, Z.~Liu, and C.~Li, ``Video instruction tuning with synthetic data,'' \emph{arXiv preprint arXiv:2410.02713}, 2024.

\bibitem{liu2024improved}
H.~Liu, C.~Li, Y.~Li, and Y.~J. Lee, ``Improved baselines with visual instruction tuning,'' in \emph{Proceedings of the IEEE/CVF Conference on Computer Vision and Pattern Recognition}, 2024, pp. 26\,296--26\,306.

\bibitem{zhu2023minigpt}
D.~Zhu, J.~Chen, X.~Shen, X.~Li, and M.~Elhoseiny, ``Minigpt-4: Enhancing vision-language understanding with advanced large language models,'' \emph{arXiv preprint arXiv:2304.10592}, 2023.

\bibitem{li2025llama}
Y.~Li, C.~Wang, and J.~Jia, ``Llama-vid: An image is worth 2 tokens in large language models,'' in \emph{European Conference on Computer Vision}.\hskip 1em plus 0.5em minus 0.4em\relax Springer, 2025, pp. 323--340.

\bibitem{ataallah2024minigpt4}
K.~Ataallah, X.~Shen, E.~Abdelrahman, E.~Sleiman, D.~Zhu, J.~Ding, and M.~Elhoseiny, ``Minigpt4-video: Advancing multimodal llms for video understanding with interleaved visual-textual tokens,'' \emph{arXiv preprint arXiv:2404.03413}, 2024.

\bibitem{liu2024llava}
H.~Liu, C.~Li, Y.~Li, B.~Li, Y.~Zhang, S.~Shen, and Y.~J. Lee, ``Llava-next: Improved reasoning, ocr, and world knowledge,'' 2024.

\bibitem{rossetto2019v3c}
L.~Rossetto, H.~Schuldt, G.~Awad, and A.~A. Butt, ``V3c--a research video collection,'' in \emph{International Conference on Multimedia Modeling}.\hskip 1em plus 0.5em minus 0.4em\relax Springer, 2019, pp. 349--360.

\bibitem{1178722}
A.~F. Smeaton, P.~Over, and W.~Kraaij, ``Evaluation campaigns and trecvid,'' in \emph{{MIR} '06: {P}roceedings of the 8th {ACM} {I}nternational {W}orkshop on {M}ultimedia {I}nformation {R}etrieval}.\hskip 1em plus 0.5em minus 0.4em\relax New York, NY, USA: ACM Press, 2006, pp. 321--330.

\end{thebibliography}
\end{document}